\documentclass{article}

\usepackage{arxiv}

\usepackage[utf8]{inputenc} 
\usepackage[T1]{fontenc}    
\usepackage{hyperref}       
\usepackage{url}            
\usepackage{booktabs}       
\usepackage{amsfonts}       
\usepackage{nicefrac}       
\usepackage{microtype}      
\usepackage{color}
\usepackage{bm}             
\usepackage[ruled,linesnumbered]{algorithm2e}
\usepackage{graphicx}
\usepackage{subfigure}
\graphicspath{{figures/}}

\definecolor{lblue}{rgb}{0.1,0.45,0.8}

\title{BikNN: Anomaly Estimation in Bilateral Domains with k-Nearest Neighbors}

\author{
  Zhongping Ji\thanks{jzp@hdu.edu.cn} \\
  Computer Science and Technology\\
  Hangzhou Dianzi University, China\\
}

\begin{document}
\maketitle

\begin{abstract}

In this paper, a novel framework for anomaly estimation is proposed.
The basic idea behind our method is to reduce the data into a two-dimensional space and then rank each data point in the reduced space.
We attempt to estimate the degree of anomaly in both spatial and density domains.
Specifically, we transform the data points into a density space and measure the distances in density domain between each point and its k-Nearest Neighbors in spatial domain.
Then, an anomaly coordinate system is built by collecting two unilateral anomalies from k-nearest neighbors of each point.
Further more, we introduce two schemes to model their correlation and combine them to get the final anomaly score.
Experiments performed on the synthetic and real world datasets demonstrate that the proposed method performs well and achieve highest average performance.
We also show that the proposed method can provide visualization and classification of the anomalies in a simple manner.
Due to the complexity of the anomaly, none of the existing methods can perform best on all benchmark datasets.
Our method takes into account both the spatial domain and the density domain and can be adapted to different datasets by adjusting a few parameters manually.

\end{abstract}


\section{Introduction}

Generally speaking, outliers, or called anomalies, are the minority of data points that have different characteristics from the normal instances. The existence of outliers is very common in data science and machine learning. The outliers often provide significant information in various application domains, such as the detection of network unauthorized access, fraud in credit card transactions and new pattern discovery. Outlier detection has been applied in the medical domain as well by monitoring vital functions of patients and it is also used for detecting failures in complex systems. Outlier can also signally impact the performance of statistical models. It thus becomes imperative to identify outliers in order to reveal new discoveries or to remove outliers from normal instances in order to maintain the performance machine learning  models.

Numerous methods have been proposed, and they detect the outliers in different ways.
We list some typical methods, such as linear models \cite{RousseeuwD99,schoelkopf2001}, distance-based methods \cite{RamRasShi00}, binary space partioning based methods \cite{Liu2008}, histogram-based methods \cite{Goldstein2012}, probabilistic methods \cite{li2020copod}, ensemble-based methods \cite{Lazarevic,Pevny16,zhao2019lscp} and neural network based methods \cite{Christopher,Liu2020}.

We briefly introduce the methods related to our method. Outlier detector with minimum covariance determinant (MCD) provides a robust estimator of the mean and covariance matrix of observations in a way that tries to minimize the influence of outliers. It finds a subset of the observations whose covariance matrix has the lowest determinant. However, as a linear model, it is not suitable for the dataset where there are multiple clusters. The outlier detector with k-Nearest Neighbors (kNN) identifies outliers by exploiting the distances between the points among neighborhoods of data points. The farther away a point is from its neighbors, the more possible the point is an outlier. Histogram-based outlier detector assumes that each dimension is independent and divides a certain amount of intervals in each dimension. The anomaly score is estimated by aggregating the density at each interval. It focuses on detecting global outliers efficiently but it might perform poor on local outlier detection. These methods identify outliers in their own ways, and each of them has its advantages and limitations.

We propose a method that combines spatial distance and probability density to reveal anomalies jointly, aiming to provide simple interpretability and even further classification and visualization of anomalies.
The main contributions of the paper can be summarized as follows:

\begin{itemize}
\item We propose a bilateral anomaly estimation by carefully combining the probability density with the k-Nearest Neighbors. To facilitate the estimation of the anomaly from the density domain, we introduce a density space and transform the data points into this space.
\item We highlight that the proposed method may provide more functions than traditional methods, such as simple interpretability, classification and visualization of the anomalies. The interpretability simply attributes the anomaly to spatial distance or probability density distance. In addition, our method can classify the outliers according to the bilateral anomalies.
\item We present a framework based on k-Nearest Neighbors for anomaly estimation, where kNN not only provides the spatial proximity, but also a local neighborhood.
    By exploring more relationships between points in the neighborhood, more unilateral anomalies can be integrated into the framework.
\end{itemize}

\section{Methodology}

We describe the main motivation of our method in the first part, and present the details of the algorithm in the following parts.

\subsection{Motivation}

For high-dimensional dataset, anomaly estimation can be regarded as a special dimensionality reduction problem.
Different from the general dimensionality reduction problem, our task does not aim to maintain the intrinsic dimension of the underlying manifold or the local structure in the original dimensions, but only focuses on maintaining the degree of anomaly.

The traditional anomaly estimators rate each data point, which is equivalent to reducing the data to one-dimensional.
We propose a bilateral scheme to estimate the anomaly of each point.
Our method consists of two main steps.
First, we reduce the data into a two-dimensional space by estimating two unilateral anomalies based on the k-nearest neighbors.
In the reduced space, most normal instances are clustered together, while the outliers scattered and the number of them tends to be small.
And then we model the correlation between the anomalies to rate the data points jointly in the reduced space.

Given a data set with $n$ data points of $d$ dimension, $\{x_k\}$, $k = 1,\cdots,n$, $x_k \in \mathbb{R}^{d}$, we need to quantify the degree of anomaly for each point.
The anomaly can be understood as the degree of deviation from most normal points.
Points that deviate farther from most points tend to be sparser.
Given a data point, we can measure its sparsity by calculating the distances between it and k-nearest neighboring points.
A point with a high sparsity value can be considered to be quite different from the surrounding points, and thus can be regarded as an outlier.
The key step of this type of methods is to evaluate the distance between two points.
In brief, our anomaly estimation depends not only on the distance in spatial domain, but also on the difference in probability density domain.

\subsection{Density-aware Distance}

The anomaly of the point is often related to the distribution of the data, so it is necessary to take the distribution into account when estimating the distance between points.
On the one hand the closer points in Euclidean space are more likely to be similar, on the other hand the denser between two points are, the less similar they are.
The denser here indicates there are more other points between two points.
This is the our motivation for introducing a density-aware distance between points for anomaly estimation.

Given two values $x_1$ and $x_2$ of a real-valued random variable $X$, a density-aware distance between them is defined by

$$
\mathbf{d}(x_1,x_2) = F_{X}(x_2)-F_{X}(x_1) = \mathrm{P}(x_1<X \leq x_2),
$$

where $F_{X}(x)=\mathrm{P}(X \leq x)$ is the cumulative distribution function (CDF) of $X$, and $\mathrm{P}(X \leq x)$ represents the probability that the random variable $X$ takes on a value less than or equal to $x$.
To define a density-aware distance for a dataset $\{x_k\}$, we use the empirical cumulative distribution function (ECDF) which is an estimation of the underlying CDF that generated the sample.
In fact, we calculate the empirical cumulative distribution function of a multivariate dataset.
Let $X \in \mathbb{R}^d$ be a d-dimensional dataset with $n$ observations and $X^j_i$ be the $i$ th observation of the $j$ th dimension.
The $j$ th ECDF is denoted by $\hat{F}_{j}(x)$ and is defined as

$$
\hat{F}_{j}(x) = \frac{1}{n} \sum_{i=1}^{n} \mathbf{I}(X^{j}_{i} \leq x)
$$

where $\mathbf{I}(\cdot)$ is the indicator function which has 2 possible values: 1 if $x_{i} \leq x$ is true, and 0 if not.
The result is a step function that increases by $\frac{1}{n}$ at each data point.

We assume that the features of data are independent of each other.
Given two data points $x_{i_1}$ and $x_{i_2}$, a density-aware distance between them is defined by

$$
\mathcal{D}(x_{i_1},x_{i_2}) = ||\mathbf{D}(x_{i_1},x_{i_2})||_{p} = ||\left<\mathbf{D}_1(x_{i_1},x_{i_2}),\mathbf{D}_2(x_{i_1},x_{i_2}),\cdots,\mathbf{D}_d(x_{i_1},x_{i_2})\right>||_{p}
$$

where the $j$ th entry of vector $\mathbf{D}(x_{i_1},x_{i_2})$ is $\mathrm{P}_j(x_{i_1,j}<X \leq x_{i_2,j})$ for $j = 1,2,\cdots,d$, and $p$ is the $p$-norm of a vector.

We first transform the data point from the original space into ECDF space by a projection function,

$$
\mathcal{P}(x_{i}) = \left<\mathrm{P}_1(X \leq x_{i,1}),\mathrm{P}_2(X \leq x_{i,2}),\cdots,\mathrm{P}_d(X \leq x_{i,d})\right>
$$

\begin{figure}
  \centering
  \hspace{-0.1in}
  \subfigure[]{
    \label{fig:dataset_2clusters:a}
    \includegraphics[angle=0,width=2.01in]{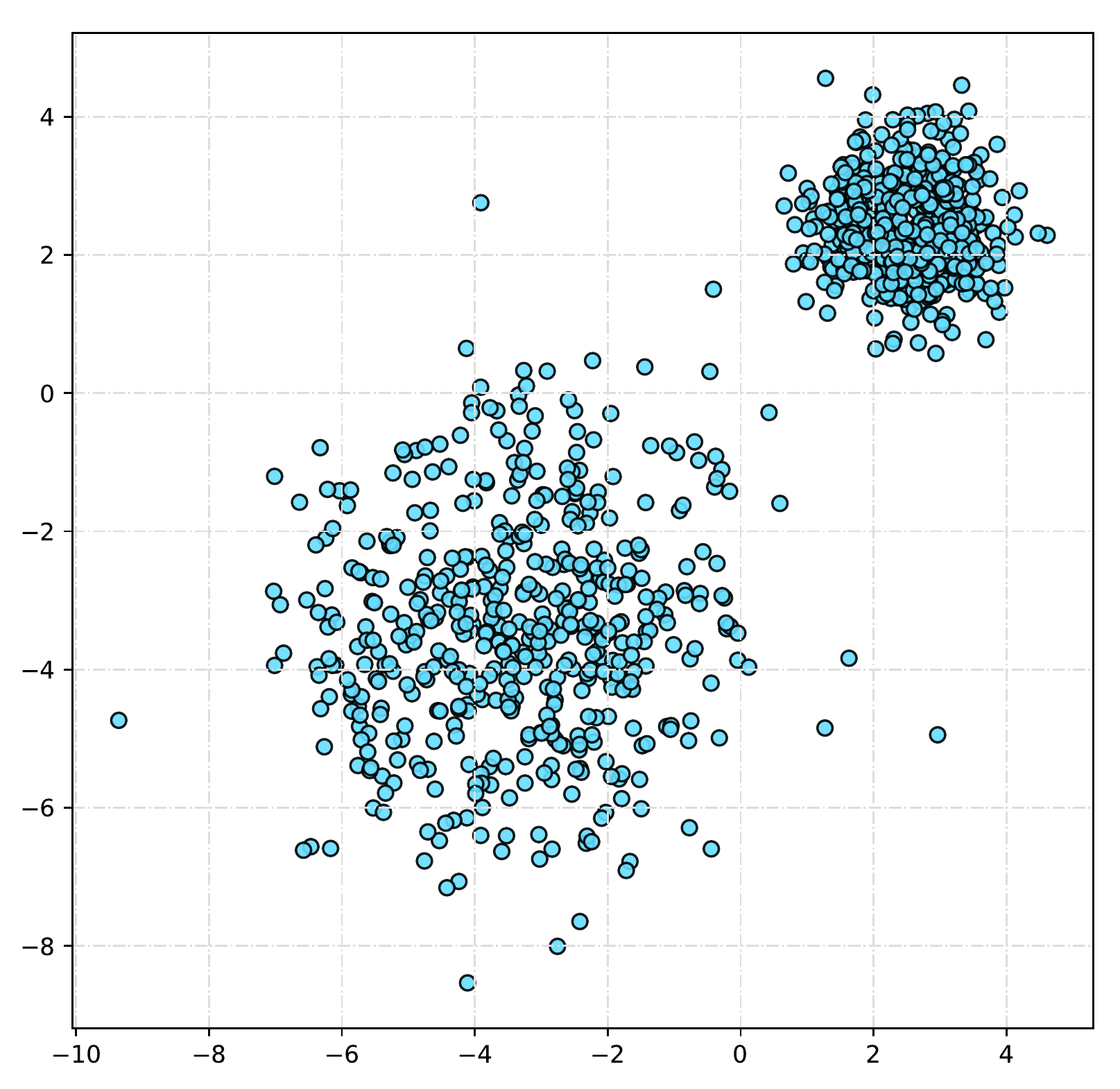}}
  \hspace{0.1in}
  \subfigure[]{
    \label{fig:dataset_2clusters:b}
    \includegraphics[angle=0,width=2.01in]{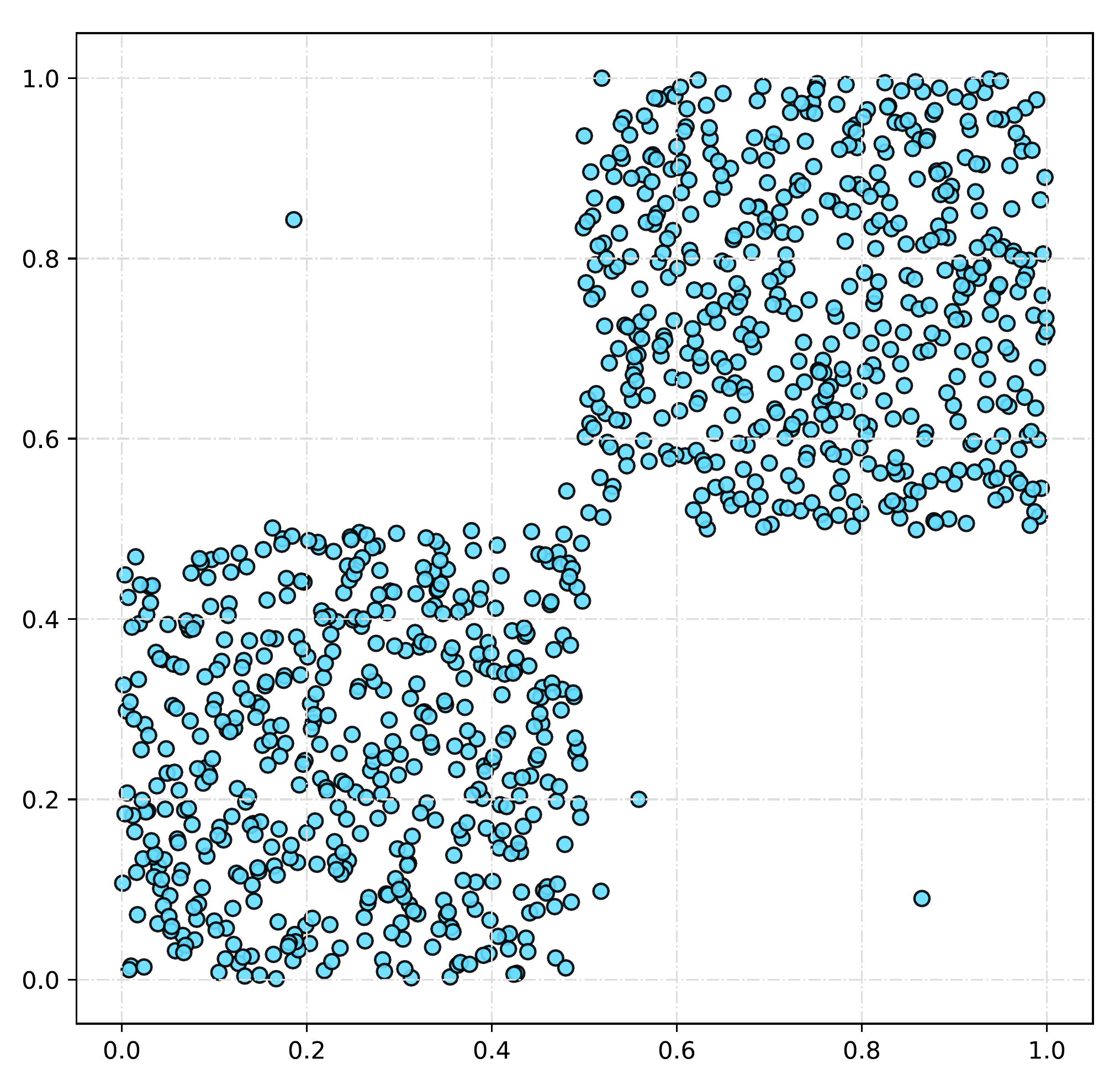}}
   \hspace{0.1in}
  \subfigure[]{
    \label{fig:dataset_2clusters:c}
    \includegraphics[angle=0,width=2.01in]{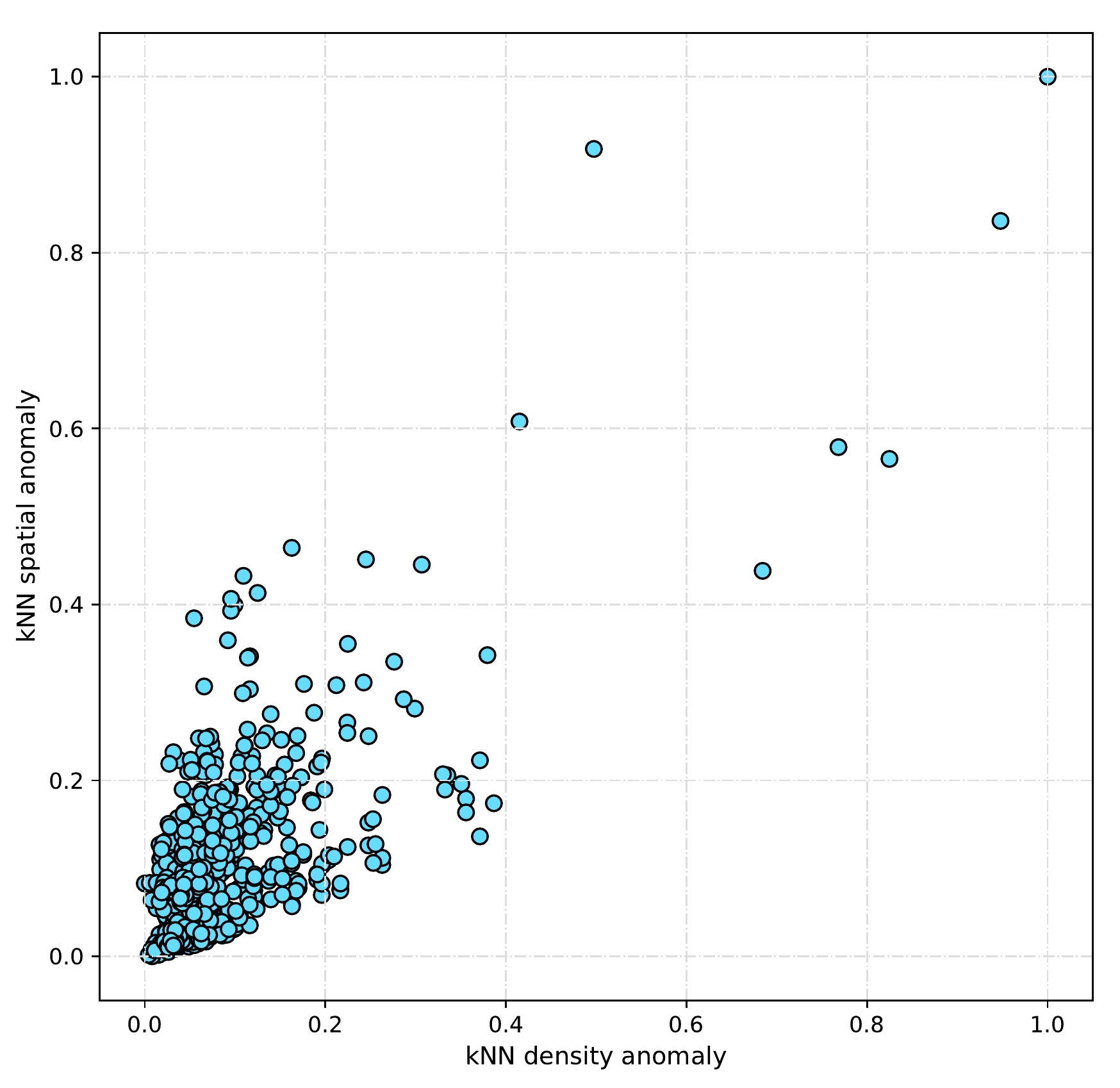}}
  \caption{\label{fig:dataset_2clusters}
Using synthetic 2D data to illustrate the ECDF space and Anomaly Space. The data points are sampled from two different Gaussian distributions. The scatter plots of
  (a) original data; (b) points transformed into ECDF space; (c) points in Anomaly Space.
}
\end{figure}

Then the density-aware distance is calculate in ECDF space in the same way as the Euclidean distance.
The original space and ECDF space of a synthetic dataset are illustrated in Figure \ref{fig:dataset_2clusters:a} and Figure \ref{fig:dataset_2clusters:b}.
It can be found from the figure that two clusters with different density in the original space are transformed into clusters with similar density in ECDF space.
In addition, two points are clearly isolated in ECDF space, which will be beneficial to outlier detection.

\subsection{2D Anomaly Space}

Now let us build the anomaly space from two unilateral distances computed in spatial domain and density domain respectively.
First, we compute the distance $\mathcal{K}_e({x}_{i})$ from $x_i$ to its $k$-th nearest neighbor point,
$$
\mathcal{K}_e({x}_{i}) = \max_{j\in N_k(i)} D_1(x_i,x_j)
$$
where $N_k(i)$ denotes the $k$ nearest neighbors of $x_i$ and $D_1(\cdot,\cdot)$ is the Minkowski distance between two points in the original space. The Euclidean distance is used by default in this paper. The function $\max$ is optional, and other functions, such as $\mbox{mean}$ or $\mbox{median}$ of the $k$ nearest neighbors can also be used. This distance can be used alone to estimate the anomaly score, and we call it \textbf{kNN spatial anomaly} in this paper.

Then, the maximum distance $\mathcal{K}_p({x}_{i})$ from $x_i$ to the $k$ nearest neighbors $N_k(i)$ is defined as,

$$
\mathcal{K}_p({x}_{i}) = \max_{j\in N_k(i)} D_2(\mathcal{P}(x_i),\mathcal{P}(x_j))
$$

where $\mathcal{P}(\cdot)$ is the projection function which transforms the data point from the original space into ECDF space and $D_2(\cdot,\cdot)$ is the Minkowski distance between two points in ECDF space. Note that we use the neighborhood of points in the original space, rather than the $k$ nearest neighbors in ECDF space. This distance also can be used alone to estimate the anomaly score, and we call it \textbf{kNN density anomaly}. The dense here is short for probability density.

These two unilateral anomalies estimate the degree of anomaly of each point from spatial position and probability density respectively. $\mathcal{K}_e({x}_{i})$ estimates the anomaly of $x_i$ in the original space, and $\mathcal{K}_p({x}_{i})$ re-estimates it in ECDF space using the same neighborhood $N_k(i)$ as $\mathcal{K}_e({x}_{i})$. As a consequence, we build a two-dimensional coordinate system for anomaly estimate, and we call it \textbf{2D Anomaly Space}. In this space, each point is represented by an ordered set of two anomalies: $(\mathcal{K}_e({x}_{i}), \mathcal{K}_p({x}_{i}))$, and two coordinate components will be combined to evaluate the anomaly score of each point.

An example is shown in Figure \ref{fig:dataset_2clusters:c}. Notice that the kNN density anomaly is plotted as a horizontal axis here.
From this figure, it can be seen that two coordinate components are not exactly on a diagonal line, indicating that there are somewhat inconsistent between two unilateral anomalies.
However, most of the points are pretty close together except for a few points on the periphery.
We need consider the correlation between the two anomalies.

\subsection{Bilateral Anomaly Estimation}

The task of anomaly estimation is to provide a ranking that reflects the degree of anomaly.
Although each unilateral anomaly mentioned above can be used to estimate the anomaly score alone, we propose two schemes to estimate the anomaly score jointly in 2D Anomaly Space.

\subsubsection{Mahalanobis Anomaly}
First, we consider the correlation between the spatial anomaly and the density anomaly.
Although there are two distinct clusters in the original dataset, when projected into the Anomaly Space, most of the data points are concentrated in the lower left corner region.

The ratio of the two coordinates $\frac{\mathcal{K}_p({x}_{i})}{\mathcal{K}_e({x}_{i})}$ indicates how distorted the density variance is relative to the spatial variance in a local neighborhood of $x_i$.
Most of points are nearly coincident, and these data points are regarded as normal instances, then the points deviating from them may be identified as anomalies.
The more a point deviates from the center of dense points off the principal direction, the more anomaly it is, so we seek an non-isotropic anomaly estimator.
The Mahalanobis distance is a common metric that can capture the non-isotropic properties of a feature space.
Thus, we estimate the Mahalanobis distance from $v(x_i)$ to the center of dense points in 2D Anomaly Space.

For a given point $x_i$, we estimates the Mahalanobis anomaly of its projected anomaly point $v(x_i)$ in Anomaly Space by
$$
\mathcal{M}_{(\bar{v},\mathbf{\Sigma})}(x_i) = \left[(v(x_i) - \bar{v})^{\top}\mathbf{\Sigma}^{-1}(v(x_i) - \bar{v})\right]^{\frac{1}{2}}
$$

where $v(x_i)=\left[\mathcal{K}_e({x}_{i}),\mathcal{K}_p({x}_{i})\right]^{\top}$, $\bar{v}$ is the center of anomaly points, and $\mathbf{\Sigma}$ is a covariance matrix estimated from the anomaly points.

To identify outliers, we estimate $\bar{v}$ and $\mathbf{\Sigma}$ from most of the normal points that are clustered together.
However, a problem arises because outliers can distort the parameters to be estimated, which makes these points appear less anomalous than they really are.
Therefore, we need a robust estimator of covariance for our task, and we use the minimum covariance determinant estimator. The idea behind it is to find a subset of normal points from which to robustly estimate $\bar{v}$ and $\mathbf{\Sigma}$.

The Mahalanobis anomaly map of the experimental dataset (see Figure \ref{fig:dataset_2clusters:c}) is shown in Figure \ref{fig:dataset_2clusters_metric:a}.
As clearly shown in the figure, the anomaly score increases much faster along the $density$-axis for this example.
It does not mean that density anomaly is more important than spatial anomaly, but the principal axis depends on specific data.

\subsubsection{Weighted Minkowski Anomaly}
\begin{figure}
  \centering
  \hspace{-0.1in}
  \subfigure[]{
    \label{fig:dataset_2clusters_metric:a}
    \includegraphics[angle=0,width=1.70in]{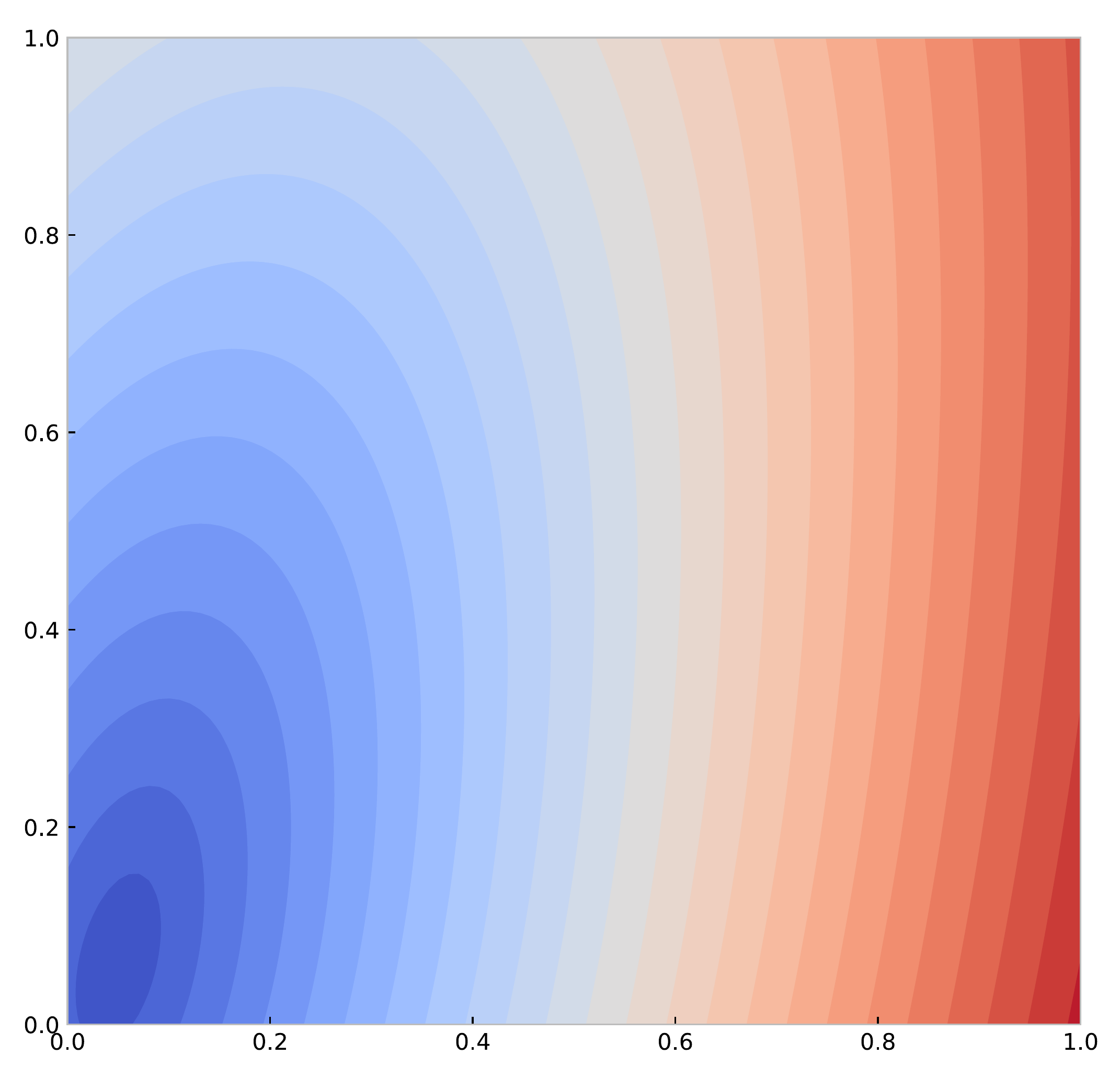}}
  \hspace{0.25in}
  \subfigure[]{
    \label{fig:dataset_2clusters_metric:b}
    \includegraphics[angle=0,width=1.70in]{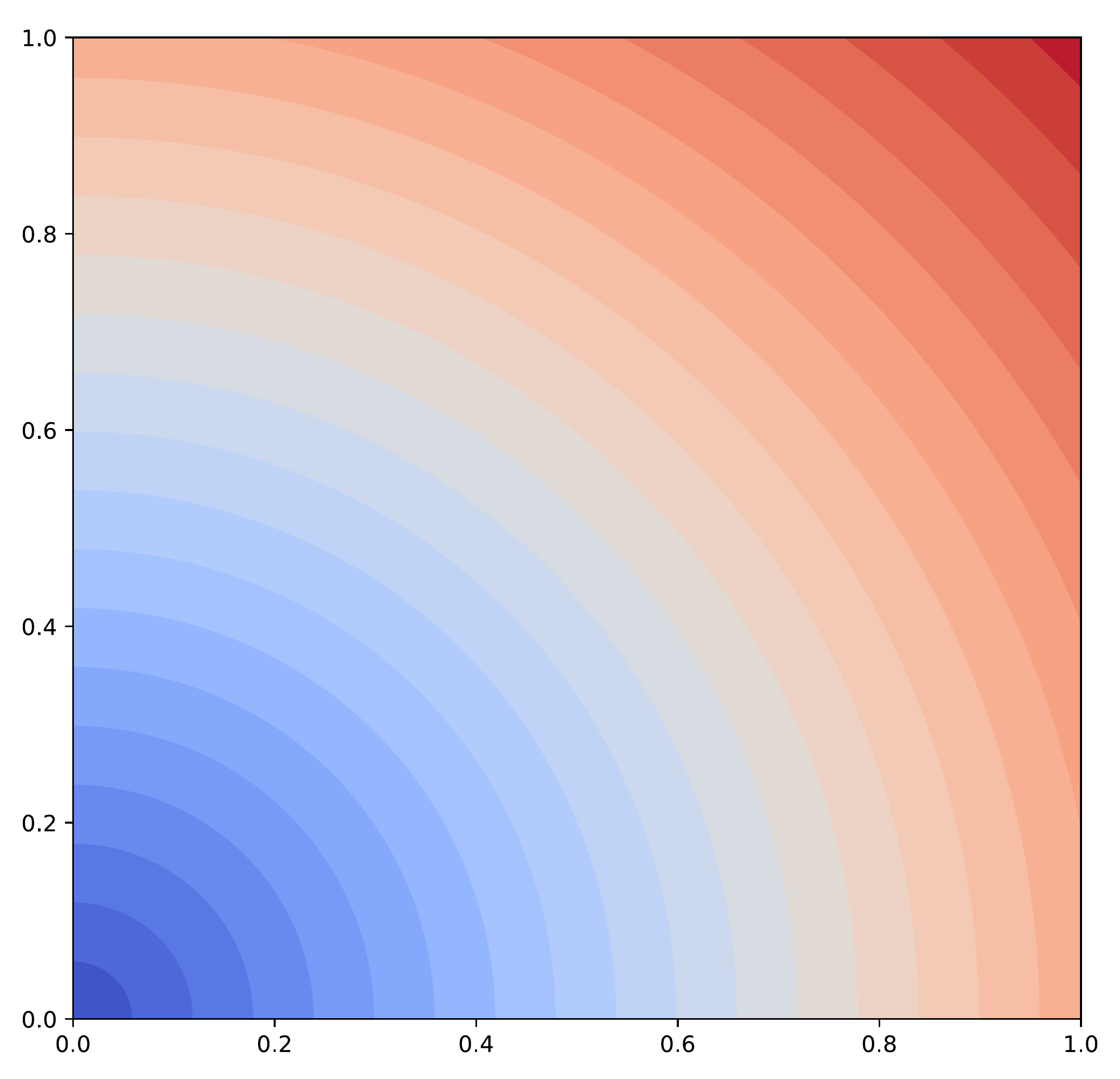}}
   \hspace{0.25in}
  \subfigure[]{
    \label{fig:dataset_2clusters_metric:c}
    \includegraphics[angle=0,width=1.70in]{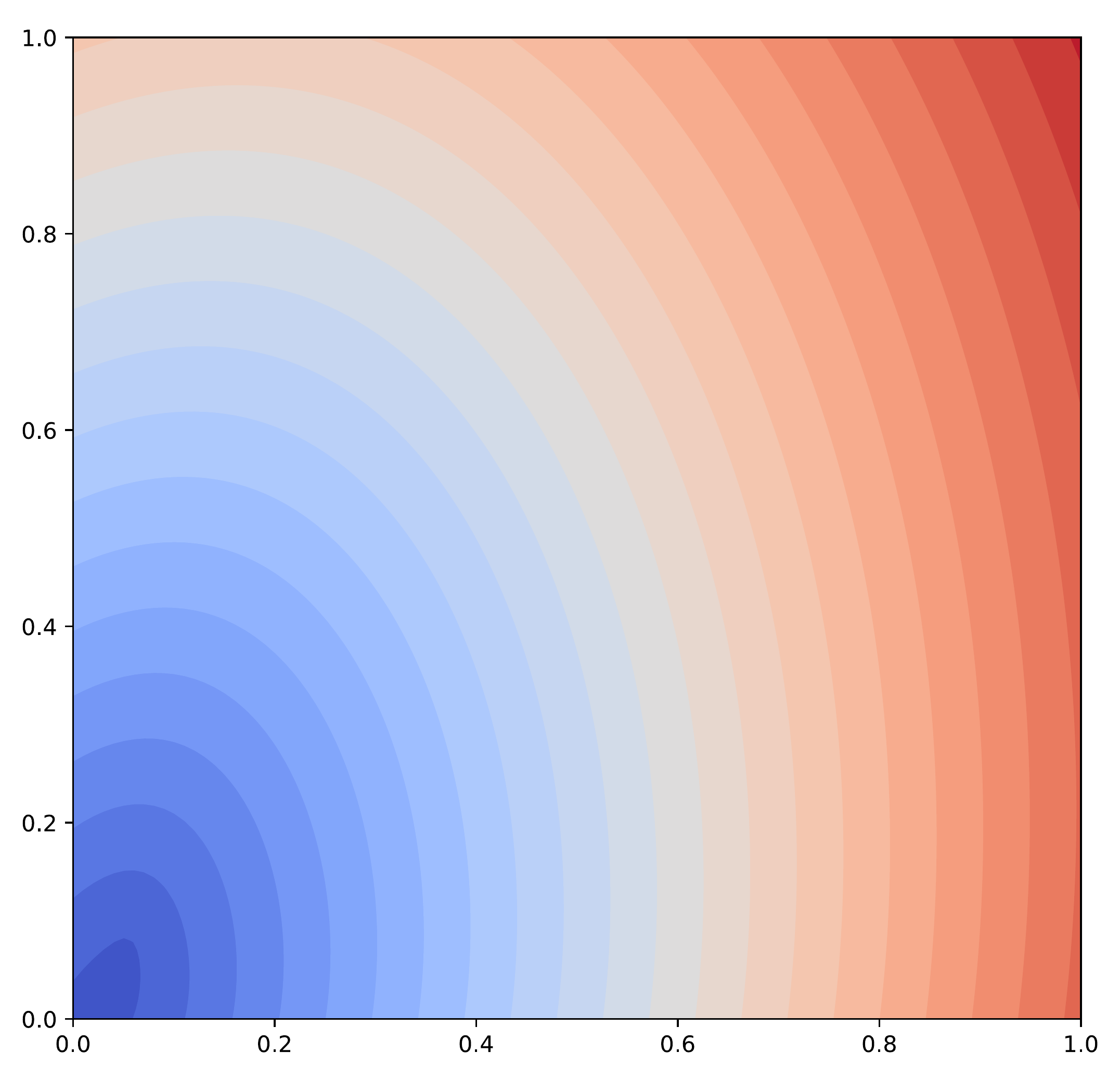}}
  \caption{\label{fig:dataset_2clusters_metric}
Illustrations of the anomaly maps of different schemes. Warmer colors indicate high anomaly values and cooler colors indicate low anomaly values. The contour plots of
  (a) Mahalanobis anomaly; (b) weighted Minkowski anomaly; (c) the combined anomaly.
}
\end{figure}

The Mahalanobis anomaly estimation can be regarded as a data-driven scheme.
In order to take advantage of two anomaly axes separately, we ignore the correlation between these two anomalies.
Specifically, the weighted Minkowski anomaly is estimated as follows:

$$
\mathcal{W}_{(w_1,w_2,p)}(x_i) =|| \tilde{v}({x}_{i})||_{p}
$$

where $\tilde{v}(x_i)=\left[w_1\mathcal{K}_e({x}_{i}),w_2\mathcal{K}_p({x}_{i})\right]^{\top}$, and $p$ is the $p$-norm of a vector.
Unlike the Mahalanobis anomaly, this scheme ignores the distribution of the anomalies.
The weighted Minkowski anomaly map of the synthetic data is shown in Figure \ref{fig:dataset_2clusters_metric:b}.

In fact, different functions or norms can be used to define the anomaly score.
In our current implementation, we use this scheme mainly for two reasons.
On the one hand, the point in the upper right region of the Anomaly Space tends to have a larger anomaly, which is not considered by the Mahalanobis scheme.
The introduction of this scheme compensates points in this region.
On the other hand, the parameters $\left[w_1,w_2 \right]$ are introduced to control the importance of the two anomalies and also to make our method include the traditional kNN methods.

\subsubsection{Anomaly Score}

The anomaly score $\mathcal{S}(x_i)$ of an instance $x_i$ is estimated by combining the Mahalanobis anomaly and the weighted Minkowski anomaly as follows,
$$
\mathcal{S}(x_i) = \mu \mathcal{M}_{(\bar{v},\mathbf{\Sigma})}(x_i) + (1-\mu)\mathcal{W}_{(w_1,w_2,p)}(x_i)
$$

This parameter $\mu \in [0,1]$ is used to balance these two anomalies.
The parameters $\bar{v}$ and $\mathbf{\Sigma}$ are detected from the data, and the parameters that need to be adjusted are mainly $w_1$, $w_2$ and $\mu$.
One of the reasons why we introduce this set of parameters is that our method becomes the traditional kNN method when we set the parameters $w_1=1$, $w_2=0$ and $\mu=1$.
The combined anomaly of the synthetic data is shown in Figure \ref{fig:dataset_2clusters_metric:c}.

\subsection{Algorithm Pseudocode}

Given a dataset, our method transforms it into 2D Anomaly Space by collecting the spatial anomaly and density anomaly from k-nearest neighbors of each point.
These two unilateral anomalies can be used to detect the outliers alone or through simple operations (such as addition and multiplication).
In order to further explore in this Anomaly Space, we introduce two schemes to compute Mahalanobis anomaly and weighted Minkowski anomaly and the final anomaly score is estimated by combining them.
We summarise the algorithm in Algorithm \ref{alg:algorithm1}.

\section{Empirical Evaluation}

This section presents the experimental results of 2 synthetic datasets and 16 real world datasets for evaluation.

\IncMargin{1.5em}
\vspace{1in}
\begin{algorithm}[H]
    \SetAlCapHSkip{.5em}
    \SetAlgoLined
    \vspace{0.05in}
	\caption{BikNN Anomaly Estimation}
	\label{alg:algorithm1}
	\KwIn{$X$ - input data}
	\KwOut{$S$ - scores of each data point in $X$}
	\BlankLine
	construct kNN for $X$, and get k-nearest neighbors $N_k(i)$ for point $x_i$;

    construct ECDF space for $X$ and projection function $\mathcal{P}(\cdot)$;

    \ForEach{each point $x_i$ in input data $X$}{
        Compute the spatial anomaly,

        $\qquad \mathcal{K}_e({x}_{i}) = \max_{j\in N_k(i)} D_1(x_i,x_j)$;

        Compute the density anomaly,

        $\qquad \mathcal{K}_p({x}_{i}) = \max_{j\in N_k(i)} D_2(\mathcal{P}(x_i),\mathcal{P}(x_j))$;
	}
    construct the 2D Anomaly Space;

    \ForEach{each point $x_i$ in input data $X$}{
        Compute the Mahalanobis anomaly,

        $\qquad \mathcal{M}_{(\bar{v},\mathbf{\Sigma})}(x_i) = \left[(v(x_i) - \bar{v})^{\top}\mathbf{\Sigma}^{-1}(v(x_i) - \bar{v})\right]^{\frac{1}{2}}$;

        Compute the weighted Minkowski anomaly,

        $\qquad \mathcal{W}_{(w_1,w_2,p)}(x_i) =||\tilde{v}({x}_{i}))||_{p}$,

        where $\tilde{v}(x_i)=\left[w_1\mathcal{K}_e({x}_{i}),w_2\mathcal{K}_p({x}_{i})\right]^{\top}$;

        Compute the anomaly score,

        $\qquad \mathcal{S}(x_i) = \mu \mathcal{M}_{(\bar{v},\mathbf{\Sigma})}(x_i) + (1-\mu)\mathcal{W}_{(w_1w_2,p)}(x_i)$;
    }
    $\mathcal{S}(X) = \left[\mathcal{S}(x_1),\mathcal{S}(x_2),\cdots,\mathcal{S}(x_n)\right]^{\top}$;
    \vspace{0.05in}
\end{algorithm}

\subsection{Bilateral Outlier Classification}

\textbf{Anomaly visualization.} Before introducing the classification, let us first describe a simple visualization of anomaly.
Our method transforms each $m$-dimensional data point $x_i$ into a 2D point $v(x_i)$. Thus, the anomaly of any high-dimensional data can be easily visualized on a plane.
Generally speaking, it is more convenient to separate outliers from normal points visually on a plane than on a single coordinate axis, since points in a higher dimensional space are apt to be separated.
In our framework, one can visualize the ordered anomaly points by plotting them on a 2D plane, and then manually mark the outliers through simple interaction.

Now we begin to introduce the classification of the outliers in our framework.
Specifically, our goal now is to classify outliers according to their anomaly coordinates in 2D Anomaly Space.
First, we extract outliers according to thresholds in spatial anomaly coordinates and in density anomaly coordinates respectively.
Imagine that two perpendicular lines divide the plane into four regions.
Except for the region where the normal data points are located, we simply classify the points in the remaining three regions into three types.
Points of type I are those with high spatial anomalies and high density anomalies, points of type II with high spatial anomalies and low density anomalies, and points of type III with low spatial anomalies and high density anomalies.

\begin{figure*}[t]
\hspace{-0.18in}
\centering
\subfigure[]{
    \begin{minipage}[b]{0.25\linewidth}
        \centering
        \label{fig:dataset_classification:a}
        \includegraphics[height=1.593in]{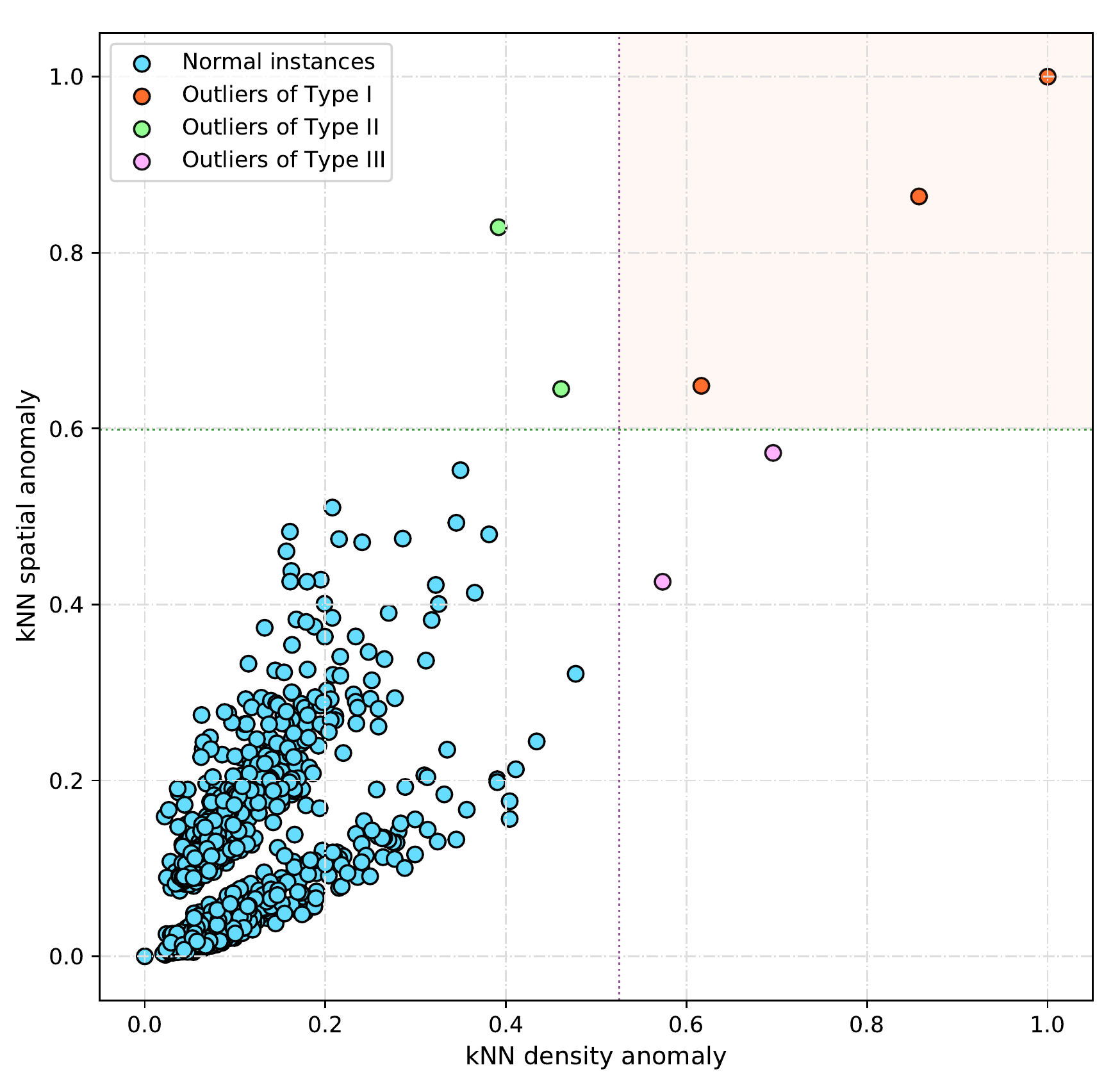}\\
        \vspace{0.0cm}
    \end{minipage}%
}%
\subfigure[]{
    \begin{minipage}[b]{0.25\linewidth}
        \centering
        \label{fig:dataset_classification:b}
        \includegraphics[height=1.55in]{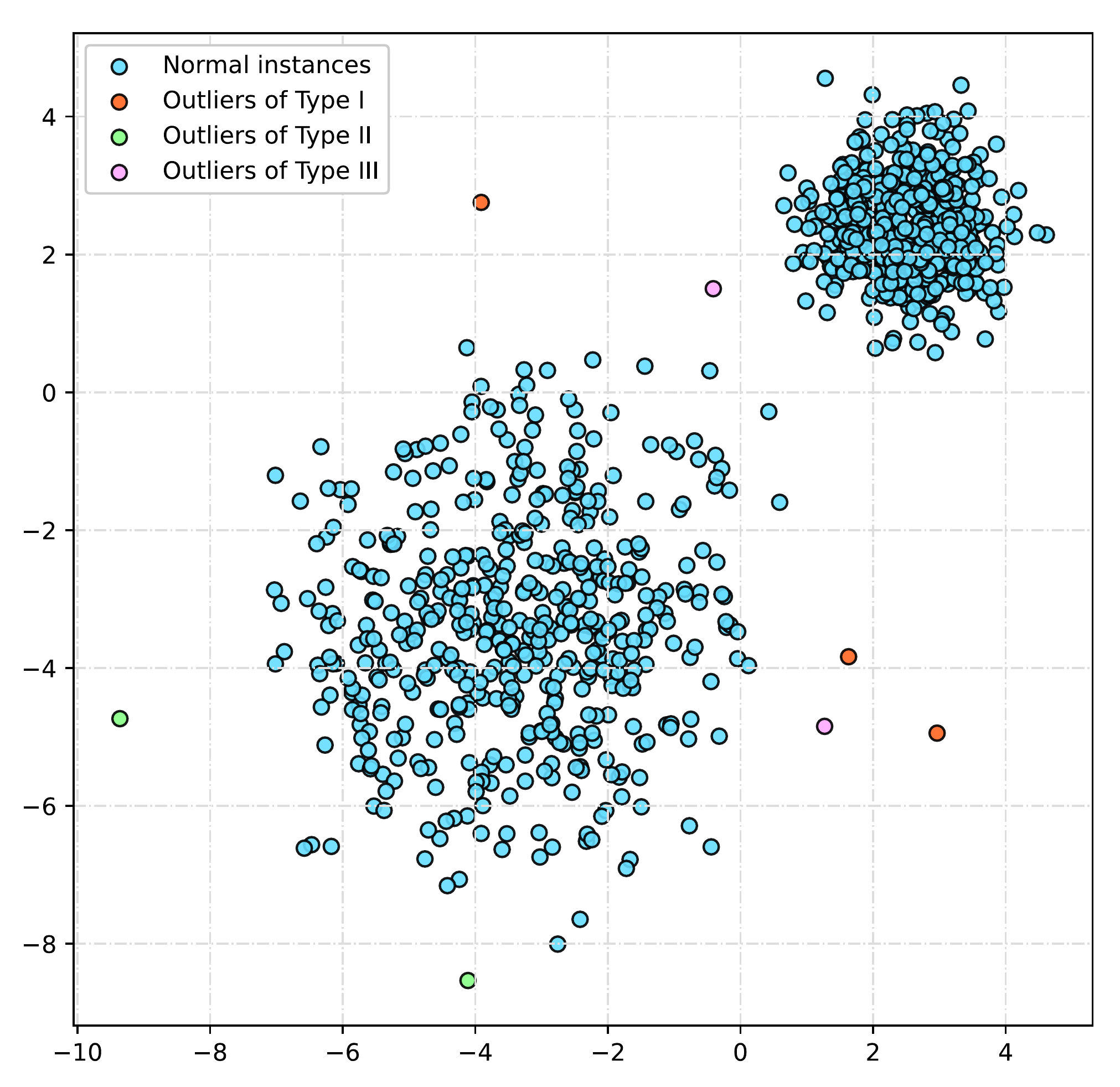}\\
        \vspace{0.11cm}
    \end{minipage}%
}%
\subfigure[]{
    \begin{minipage}[b]{0.25\linewidth}
        \centering
        \label{fig:dataset_classification:c}
        \includegraphics[height=1.55in]{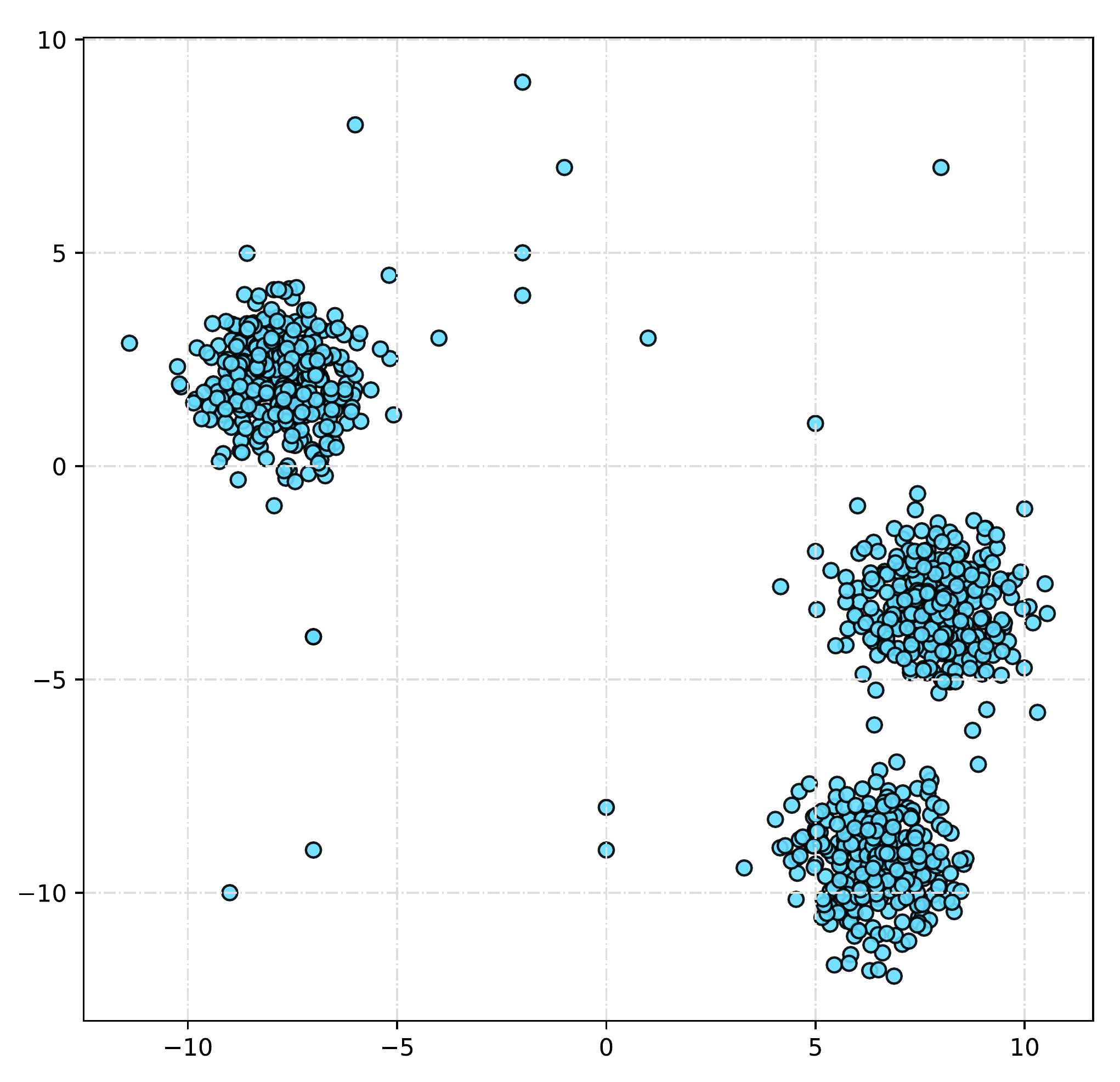}\\
        \vspace{0.11cm}
    \end{minipage}%
}%
\subfigure[]{
    \begin{minipage}[b]{0.25\linewidth}
        \centering
        \label{fig:dataset_classification:d}
        \includegraphics[height=1.593in]{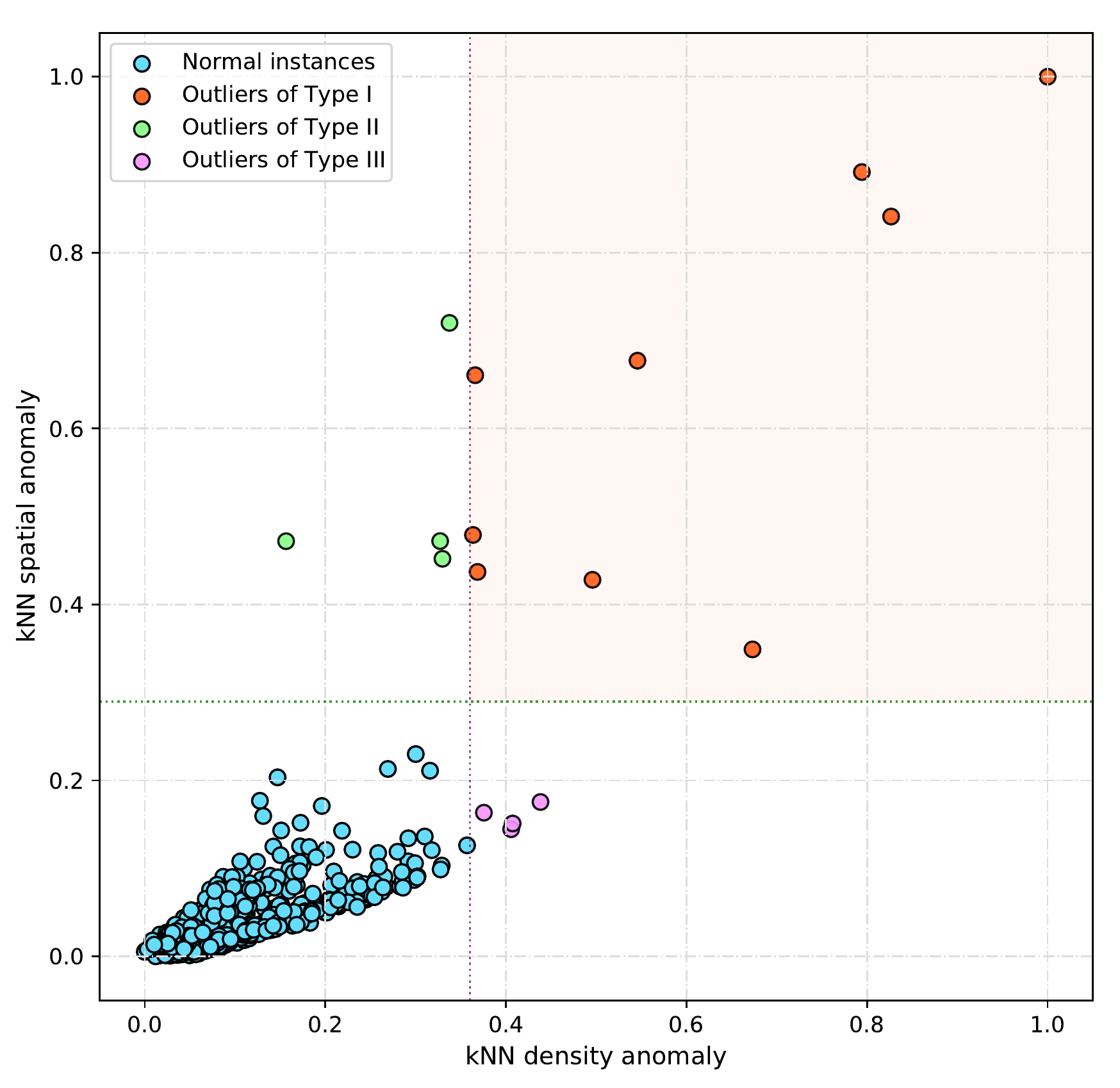}\\
        \vspace{0.0cm}
    \end{minipage}%
}%
\centering
\caption{Using the spatial anomaly and density anomaly to classify the outliers. We use $k=30$ nearest neighbors in these examples and color the outliers according to their types.
  (a) bilateral classifying the outliers in Anomaly Space; (b) scatter plot in the original space; (c) a synthetic data with three clusters; (d) bilateral classifying the outliers of (c) in Anomaly Space.}
\vspace{0.0cm}
\label{fig:classification}
\end{figure*}

An illustration of bilateral classifying the outliers in Anomaly Space is given in Figure \ref{fig:dataset_classification:a}. The thresholds for both coordinate components are determined by the number of outliers which is given in advance. In this example, we assume that the number of outliers is $5$, which results in $3$ outliers of type I, $2$ outliers of type II and $2$ outliers of type III.
In order to observe the positions of outliers with different types in the original space, we colored the points according to their types in Figure \ref{fig:dataset_classification:b}.
In order to explain the difference between three types, it is assumed that all outliers can belong to one or some adjacent clusters.
For the points of type I, they look far away from the dense points in the spatial domain, but they tend to belong to different clusters according to different features.
For example, the red point in the upper region of Figure \ref{fig:dataset_classification:b} belongs to the cluster below classified by its $x$-coordinate, and to the cluster above by its $y$-coordinate.
For the points of type II, they are also far away dense points, but clearly belong to the nearest cluster.
And the points of type III lie among different clusters, with a moderately sparse neighborhood.
The second example with three clusters in the original space is shown in Figure \ref{fig:dataset_classification:c}.
The number of outliers for this example is set to $13$ and we obtain $9$ outliers of type I.

From these two examples, it can be seen that the points may be scattered in multiple clusters in the original space, but most of the points are clustered in the lower left corner region in the anomaly space (see the Figure \ref{fig:dataset_classification:a} and Figure \ref{fig:dataset_classification:d}).
Note that the outliers classified here are not the final results, but the explanatory ranks obtained by observing the data points from a unilateral anomaly coordinate.
This kind of ranks can provide useful guidance for ones to further investigate what features make certain points potentially outliers.

\subsection{Comparison}

\subsubsection{Synthetic Dataset}

\begin{figure}
  \centering
  \subfigure{
    \label{fig:contour_1:a}
    \includegraphics[angle=0,width=6.5in]{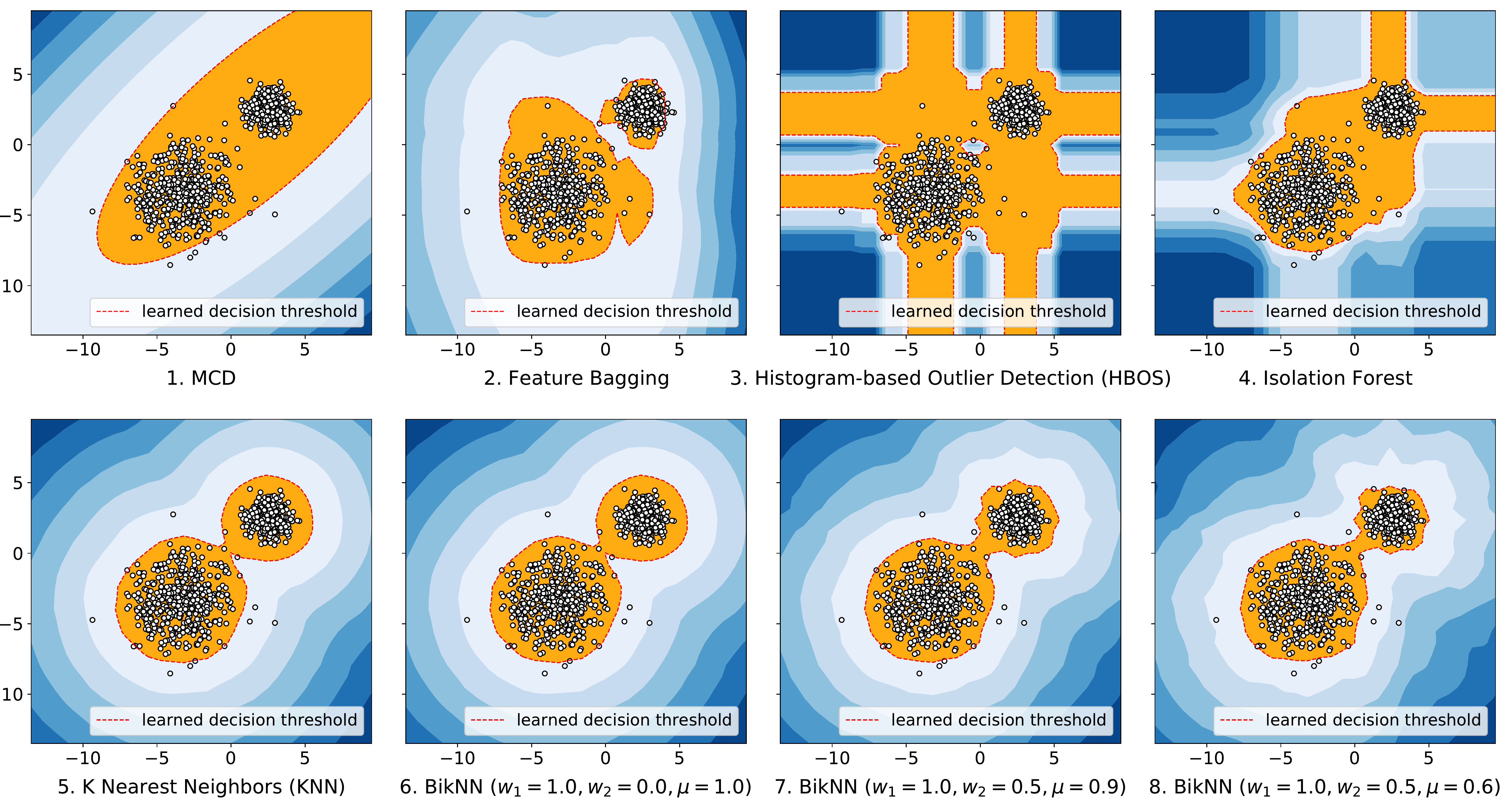}}
  \caption{\label{fig:contour_1}
Contour plots of different prediction models. The red dashed line in each subplot represents the threshold which is used to predict binary outlier labels.
The region where the normal points are located is highlighted in orange.
}
\end{figure}

First, we compare model performance on outlier detection with the synthetic datasets.
A comparison of contour plots derived from the prediction functions of different models is shown in Figure \ref{fig:contour_1}.
In this example, we define the number of outliers as $10$.
The threshold is determined by the scores of the first $n\!-\!10$ data points which are sorted by score in ascending order, and the last $10$ data points are identified as outliers.
The anomaly scores of input samples are estimated using different models.
For consistency, instances with larger anomaly scores are identified as outliers for all models.

As can be seen from the figure, the contours of different methods have different shapes.
The MCD model is to be applied on Gaussian-distributed data or on data drawn from a unimodal, symmetric distribution.
In this example, the contour of this model has a large deviation from the distribution of the data.
Feature Bagging is an ensemble-based model.
It aggregates the scores of multiple outlier detectors which are applied using randomly selected subset of features.
This means a diverse set of estimators is created by introducing randomness in the base detector construction. This also introduces randomness to the model, so the result of each run will be slightly different.
The contour of HBOS consists of many rectangles oriented along the coordinate axis.
This method divides several intervals in each dimension and the anomaly score is estimated by aggregating the density at each interval.
It focuses on detecting global outliers efficiently but it might perform poor on local outlier detection.
HBOS divides the space implicitly by probability density, while Isolation Forest divides the space explicitly by constructing some isolation trees.
To recursively divide the feature space, the isolation tree is built by randomly selecting the attribute and the split value.
For this example, Isolation Forest produces contours somewhat similar to HBOS.

The proposed method combines spatial distance and probability density, so it may resemble the contours of these two types of methods.
We know that the introduction of parameters $w_1$ and $w_2$ is to adjust the importance of the spatial domain and density domain respectively.
The larger the parameter $w_2$, the more the result is biased towards the contour derived from the density anomaly  (see the 6th and 7th figures in Figure \ref{fig:contour_1}).
In addition, the contour of our method will slightly vary with different values of parameter $\mu$.
When the parameters $w_1$ and $w_2$ are fixed, the larger the parameter $\mu$, the smoother the contour, and the smaller the parameter $\mu$, the more the contour is affected by data points near the periphery.
When we set the parameters $w_1=1$, $w_2=0$ and $\mu=1$, our method becomes the kNN detector (see the 5th and 6th figures in Figure \ref{fig:contour_1}).
By controlling the contours through few parameters, our method can identify different types of outlier points.

\subsubsection{Benchmark Datasets}

Now we start to compare the performance of BikNN with other outlier detectors on a set of benchmark datasets.
We select a set of various outlier detection models for comparison.
These models include Feature Bagging (FB), Histogram-based Outlier Score (HBOS),
Isolation Forest (IForest), Lightweight On-line Detector of Anomalies (LODA),
Local Outlier Factor(LOF), One-Class Support Vector Machines (OCSVM), Locally Selective Combination of Parallel Outlier Ensembles (LSCP), Copula-Based Outlier Detection (COPOD) and k Nearest Neighbors (kNN).
The implementations of these detectors can be found in PyOD \cite{zhao2019pyod}.
We implement our method in Python and compare it with other outlier detection models on a PC with an Intel(R) Core(TM) i7-7800X @ 3.5GHz and 8GB of memory.

\begin{figure}
  \centering
  \hspace{-0.1in}
  \subfigure[]{
    \label{fig:realdata_xy2d:a}
    \includegraphics[angle=0,width=1.5396in]{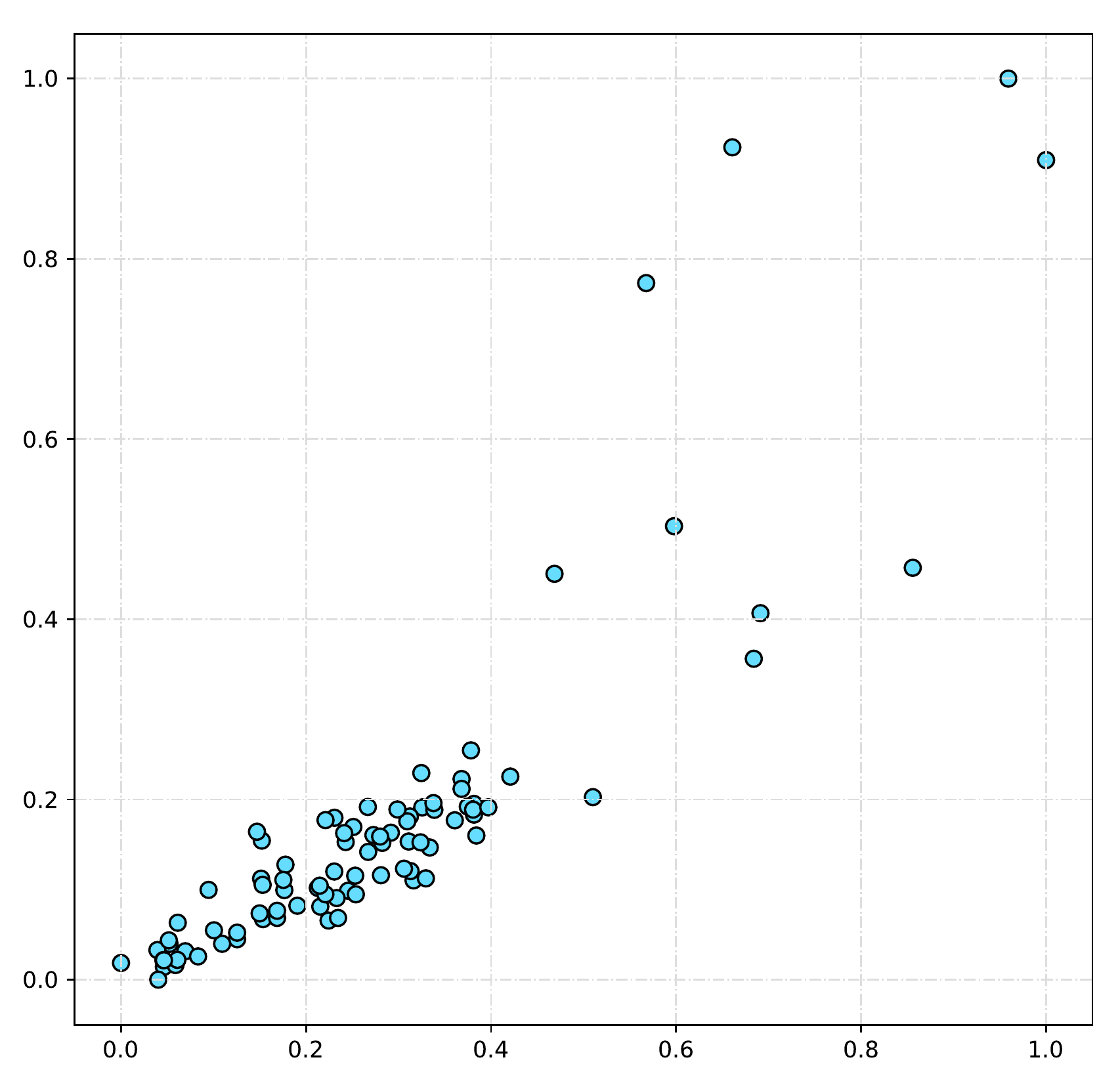}}
  \hspace{0.02in}
  \subfigure[]{
    \label{fig:realdata_xy2d:b}
    \includegraphics[angle=0,width=1.5396in]{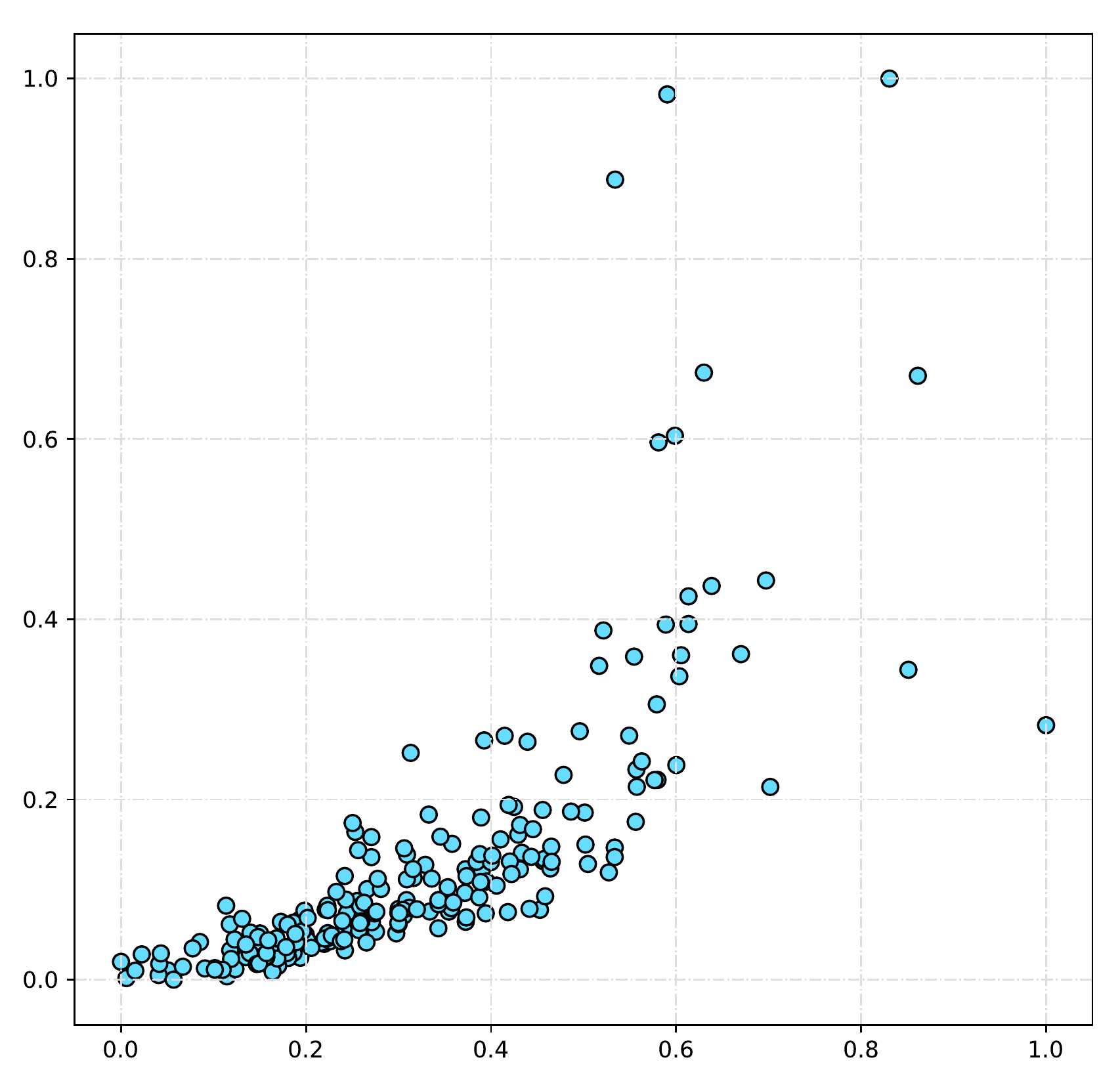}}
   \hspace{0.02in}
  \subfigure[]{
    \label{fig:realdata_xy2d:c}
    \includegraphics[angle=0,width=1.5396in]{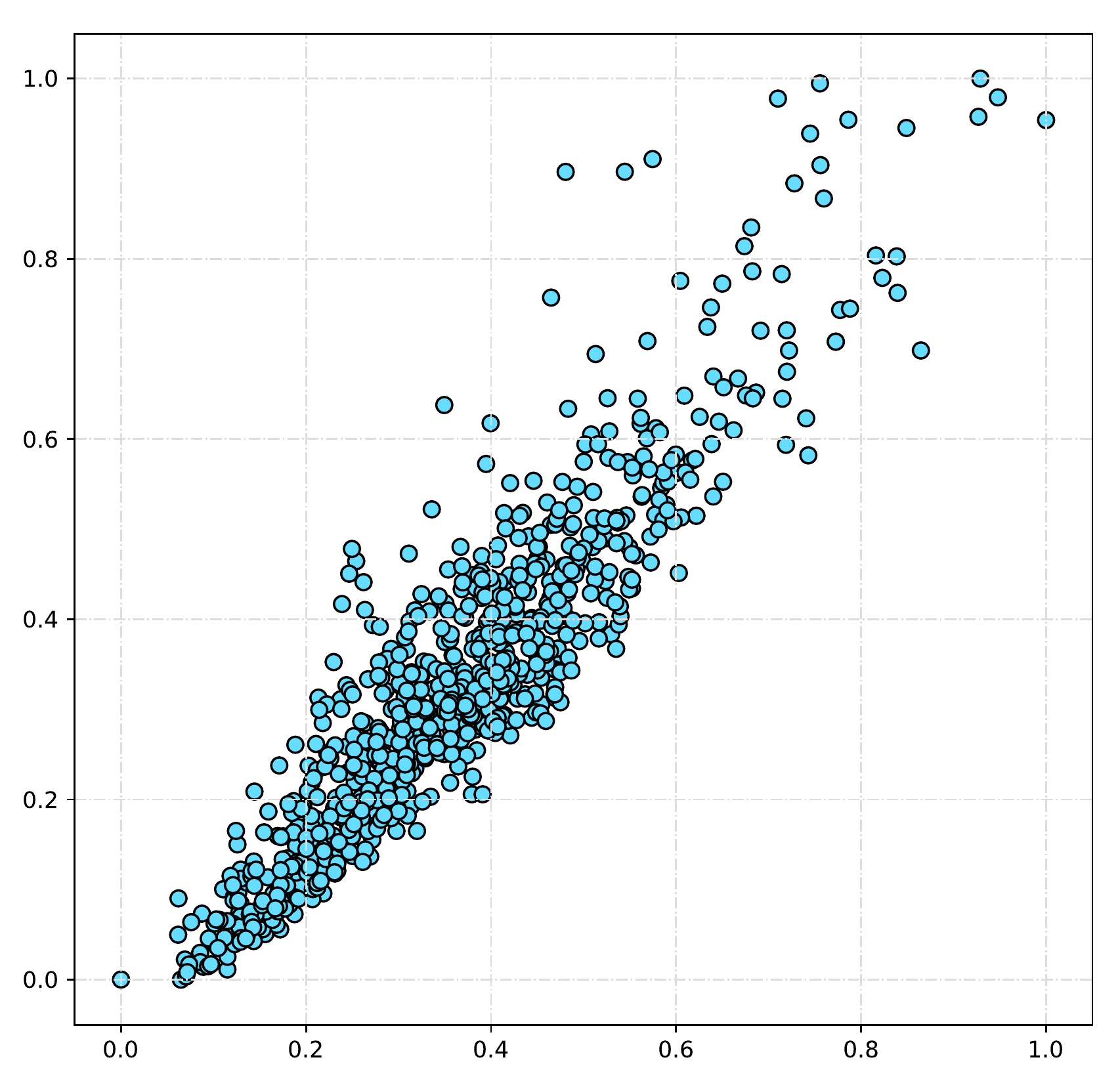}}
   \hspace{0.02in}
  \subfigure[]{
    \label{fig:realdata_xy2d:d}
    \includegraphics[angle=0,width=1.5396in]{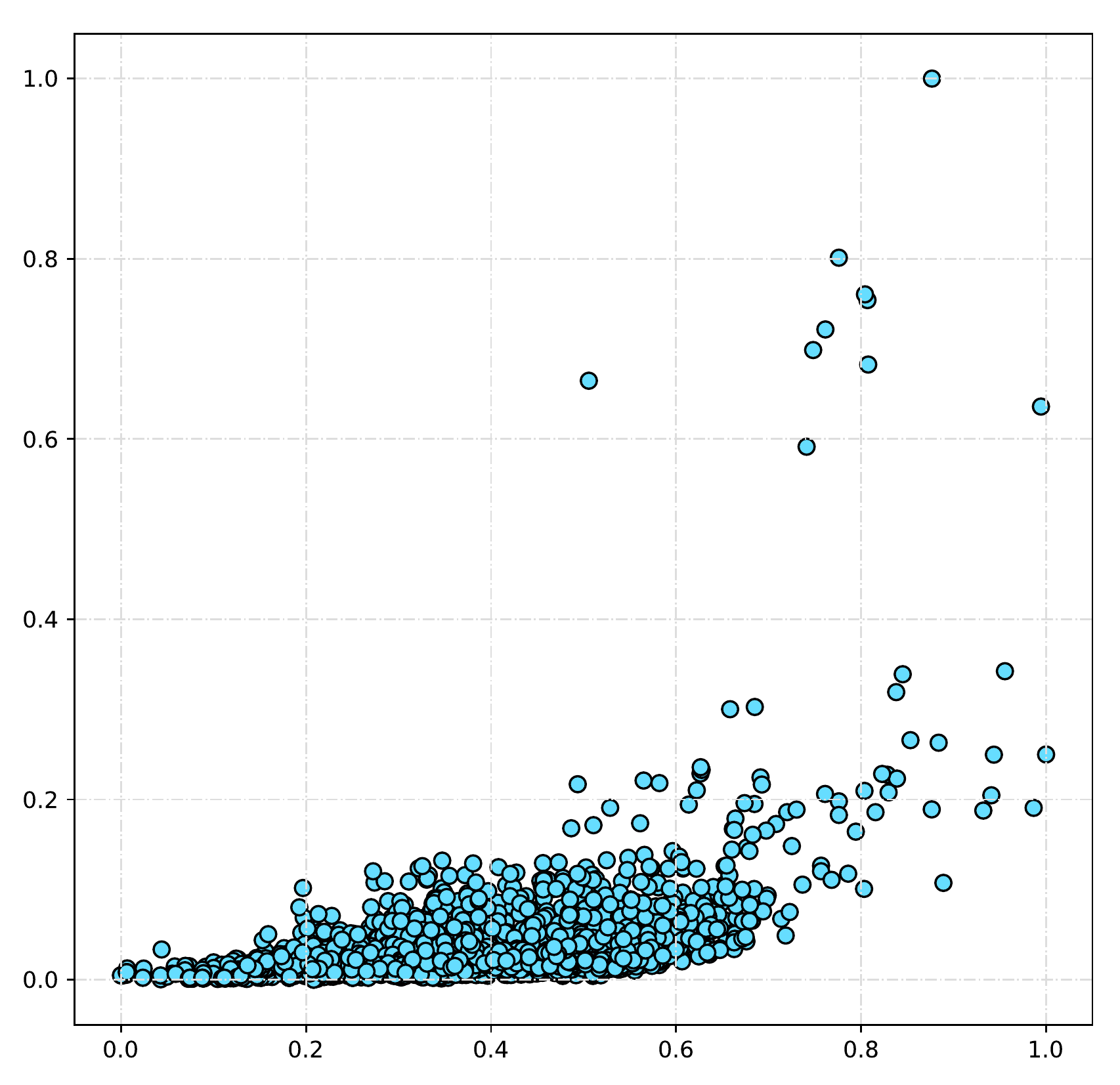}}
  \caption{\label{fig:realdata_xy2d}
Scatter plots of the anomaly maps: corresponding to the datasets lympho (\#148), wbc (\#378), vowels (\#1456) and annthyroid (\#7200) from left to right.
}
\end{figure}

In this study, we use 16 public benchmark datasets from ODDS \cite{Rayana2016} to evaluate different detectors.
Scatter plots of several anomaly maps with different numbers of data points are illustrated in Figure \ref{fig:realdata_xy2d}.
The scattered points shown here are the points in the training set of each data set.
Area under the receiver operating characteristic (ROC-AUC, shown in Table \ref{table_roc}) and average precision (AP, shown in Table \ref{table ap}) are evaluated by taking the average score of 10 independent trials.
In each trial, 60\% of the data and the remaining 40\% split from each dataset are set as training and test sets.

Our methods can achieve better performance by tuning the parameters carefully, but here we just simply set the parameters of \textbf{BikNN1} ($w_1=1,w_2=0.25,\mu=0.5$), \textbf{BikNN2} ($w_1=0.5,w_2=0.5,\mu=0.5$) and \textbf{BikNN3} ($w_1=0,w_2=1,\mu=1$).
We use $k=30$ nearest neighbors for kNN and our BikNNs in this experiment.
In fact, the kNN is a special case of our detectors, corresponding to the parameters $w_1=1,w_2=0,\mu=1$.

What is noteworthy is that \textbf{BikNN1} and \textbf{BikNN2} combine the spatial anomaly and the density anomaly to estimate the anomaly score, while \textbf{BikNN3} only uses the density anomaly.
The main purpose of \textbf{BikNN3} is to demonstrate the performance of an model that only uses the density anomaly.
Although the average score of \textbf{BikNN3} is not the highest, but it achieves the highest score on some of these datasets.
That is, for some datasets, a model estimating the anomaly only in density domain can achieve highest performance.
According to the average ROC-AUC and average AP, our model \textbf{BikNN1} performs best of all outlier detection models.

It can be seen from Table \ref{table_roc} and Table \ref{table ap} that none of these models can outperform all or most of other models on all datasets.
The reason behind may be that the distribution of anomalies varies greatly among these datasets,
and thus it is a topic worth studying to select the best model for a given dataset.
Although none of our models with fixed parameters can outperform all other models on all datasets,
our method takes into account both the spatial domain and the density domain and attempts to provide a mechanism to adapt to different datasets by adjusting a few parameters manually.

\begin{table*}[ht]
\setlength{\tabcolsep}{3pt}
\centering
\caption{ROC-AUC scores of model performance, highest score highlighted in bold black and lowest score in light blue.}
\label{table_roc}
\vspace{0.1in}
\footnotesize
\scalebox{0.78}{
\begin{tabular}{llll|cccccccc|cccc}
\toprule
      \textbf{Data} &    $n$ &   $d$ & \textbf{Outlier\%} &        \textbf{FB} &      \textbf{HBOS} &   \textbf{IForest} &      \textbf{LODA} &       \textbf{LOF} &     \textbf{OCSVM} &      \textbf{LSCP} &     \textbf{COPOD} &       \textbf{kNN} &    \textbf{BikNN1} &    \textbf{BikNN2} &    \textbf{BikNN3} \\
\midrule
annthyroid & 7200 &   6 &       7.4167 &  0.786612 &  0.621563 &  0.824550 &  \textcolor{lblue}{0.509200} &  0.724600 &  0.687100 &  0.720541 &  0.779862 &  0.753987 &  \textbf{0.825375} &  0.804138 &  0.768850 \\
arrhythmia &  452 & 274 &      14.6018 &  0.765500 &  0.808813 &  0.799238 &  \textcolor{lblue}{0.719275} &  0.767063 &  0.770212 &  0.720962 &  0.795750 &  0.778563 &  0.779075 &  0.789025 &  \textbf{0.811650} \\
   breastw &  683 &   9 &      34.9927 &  \textcolor{lblue}{0.365725} &  0.985525 &  0.989262 &  0.987762 &  0.467275 &  0.947987 &  0.378989 &  \textbf{0.994813} &  0.985962 &  0.987275 &  0.984300 &  0.925975 \\
    cardio & 1831 &  21 &       9.6122 &  0.608938 &  0.839250 &  0.921913 &  0.874938 &  \textcolor{lblue}{0.576963} &  \textbf{0.936462} &  0.668505 &  0.920750 &  0.880375 &  0.866637 &  0.817113 &  0.764538 \\
     glass &  214 &   9 &       4.2056 &  0.818138 &  0.704813 &  0.659763 &  0.501700 &  \textbf{0.849325} &  \textcolor{lblue}{0.465112} &  0.751617 &  0.638800 &  0.672400 &  0.713263 &  0.754825 &  0.795238 \\
ionosphere &  351 &  33 &      35.8974 &  0.879550 &  \textcolor{lblue}{0.548212} &  0.834375 &  0.787712 &  0.882563 &  0.836638 &  \textbf{0.910202} &  0.790050 &  0.845687 &  0.855250 &  0.845775 &  0.825038 \\
    letter & 1600 &  32 &       6.2500 &  \textbf{0.869163} &  0.575312 &  0.605075 &  \textcolor{lblue}{0.542438} &  0.858675 &  0.572338 &  0.851395 &  0.543425 &  0.775875 &  0.760375 &  0.784887 &  0.806063 \\
    lympho &  148 &  18 &       4.0541 &  0.969487 &  \textbf{0.998537} &  0.990625 &  \textcolor{lblue}{0.768062} &  0.968975 &  0.973237 &  0.985741 &  0.997975 &  0.982400 &  0.980563 &  0.980950 &  0.988437 \\
 satellite & 6435 &  36 &      31.6395 &  0.554063 &  0.757487 &  0.710675 &  0.613587 &  0.553787 &  0.673163 &  0.564086 &  0.642288 &  0.733475 &  \textbf{0.794125} &  0.758638 &  \textcolor{lblue}{0.525113} \\
satimage-2 & 5803 &  36 &       1.2235 &  0.404413 &  0.987312 &  0.996463 &  0.993962 &  \textcolor{lblue}{0.401138} &  0.998563 &  0.728120 &  0.986812 &  0.999013 &  \textbf{0.999062} &  0.998188 &  0.978550 \\
    speech & 3686 & 400 &       1.6549 &  0.478687 &  0.460688 &  0.469837 &  0.481550 &  0.481762 &  0.454275 &  0.486077 &  0.477587 &  0.467162 &  \textcolor{lblue}{0.444650} &  0.447075 &  \textbf{0.486737} \\
   thyroid & 3772 &   6 &       2.4655 &  0.849062 &  0.946525 &  \textbf{0.979088} &  0.817887 &  \textcolor{lblue}{0.801625} &  0.960925 &  0.909767 &  0.940312 &  0.967813 &  0.969400 &  0.927325 &  0.814537 \\
 vertebral &  240 &   6 &      12.5000 &  0.422762 &  \textcolor{lblue}{0.307887} &  0.386725 &  0.363712 &  0.437975 &  \textbf{0.461525} &  0.371105 &  0.335838 &  0.349263 &  0.342388 &  0.385200 &  0.433662 \\
    vowels & 1456 &  12 &       3.4341 &  0.927500 &  0.705938 &  0.764050 &  0.696287 &  0.928588 &  0.791700 &  0.938970 &  \textcolor{lblue}{0.511700} &  0.942512 &  0.931663 &  0.939362 &  \textbf{0.961137} \\
       wbc &  378 &  30 &       5.5556 &  0.946800 &  0.953750 &  0.942788 &  0.944375 &  0.945187 &  0.947400 &  0.924289 &  \textbf{0.964900} &  0.948013 &  0.947150 &  0.926162 &  \textcolor{lblue}{0.867888} \\
      wine &  129 &  13 &       7.7519 &  0.912212 &  0.910037 &  0.792338 &  0.837075 &  \textbf{0.914875} &  \textcolor{lblue}{0.689525} &  0.908111 &  0.905050 &  0.873837 &  0.837675 &  0.853975 &  0.913675 \\
\midrule
      AVG & 2149 &  58.8 &  11.4535 & 0.722413  & 0.756978  & 0.791673 & 0.714970 & 0.722523 & 0.760385 & 0.740945 & 0.764120 & 0.809771 & 0.814620 & 0.812309  & 0.791693  \\
\bottomrule
\end{tabular}
}
\end{table*}

\begin{table*}[ht]
\setlength{\tabcolsep}{3pt}
\centering
\caption{Average precision of model performance, highest score highlighted in bold black and lowest score in light blue.}
\label{table ap}
\vspace{0.1in}
\footnotesize
\scalebox{0.78}{
\begin{tabular}{llll|cccccccc|cccc}
\toprule
      \textbf{Data} &    $n$ &   $d$ & \textbf{Outlier\%} &        \textbf{FB} &      \textbf{HBOS} &   \textbf{IForest} &      \textbf{LODA} &       \textbf{LOF} &     \textbf{OCSVM} &      \textbf{LSCP} &     \textbf{COPOD} &       \textbf{kNN} &    \textbf{BikNN1} &    \textbf{BikNN2} &    \textbf{BikNN3} \\
\midrule
annthyroid & 7200 &   6 &       7.4167 &  0.209350 &  0.281313 &  \textbf{0.327875} &   \textcolor{lblue}{0.117313} &  0.221325 &  0.250625 &  0.218600 &  0.235600 &  0.288162 &  0.283525 &  0.265612 &  0.267363 \\
arrhythmia &  452 & 274 &      14.6018 &  \textcolor{lblue}{0.352950} &  \textbf{0.508750} &  0.469612 &  0.410762 &  0.360187 &  0.378675 &  0.363325 &  0.428175 &  0.387462 &  0.369800 &  0.385600 &  0.446912 \\
   breastw &  683 &   9 &      34.9927 &  \textcolor{lblue}{0.039125} &  0.941825 &  0.942538 &  0.942300 &  0.253912 &  0.899500 &  0.203487 &  \textbf{0.950650} &  0.945750 &  0.946950 &  0.937025 &  0.789212 \\
    cardio & 1831 &  21 &       9.6122 &  \textcolor{lblue}{0.150062} &  0.453025 &  0.499850 &  \textbf{0.506012} &  0.161775 &  0.515200 &  0.177325 &  0.532212 &  0.464075 &  0.473125 &  0.396275 &  0.331112 \\
     glass &  214 &   9 &       4.2056 &  0.062500 &  \textcolor{lblue}{0.000000} &  \textcolor{lblue}{0.000000} &  \textcolor{lblue}{0.000000} &  \textbf{0.177075} &  0.062500 &  0.168750 &  \textcolor{lblue}{0.000000} &  0.062500 &  0.062500 &  0.062500 &  0.125000 \\
ionosphere &  351 &  33 &      35.8974 &  0.717700 &  \textcolor{lblue}{0.335338} &  0.641600 &  0.619587 & 0.721975 &  0.720037 &  \textbf{0.760775} &  0.583350 &  0.623862 &  0.694300 &  0.691162 &  0.642500 \\
    letter & 1600 &  32 &       6.2500 & 0.352937 &  0.064525 &  0.068775 &  0.075737 &  \textbf{0.356325} &  0.105225 &  0.350912 &  \textcolor{lblue}{0.024688} &  0.201238 &  0.191513 &  0.251988 &  0.264513 \\
    lympho &  148 &  18 &       4.0541 &  0.645837 &  \textbf{0.958338} &  0.677087 &  \textcolor{lblue}{0.208338} &  0.645837 &  0.645837 &  0.756250 &  0.927088 &  0.645837 &  0.645837 &  0.645837 &  0.635412 \\
 satellite & 6435 &  36 &      31.6395 &  0.390538 &  0.575013 &  0.583037 &  0.496813 &  \textcolor{lblue}{0.384313} &  0.548038 &  0.400788 &  0.486013 &  0.532925 &  \textbf{0.625937} &  0.587238 &  0.407512 \\
satimage-2 & 5803 &  36 &       1.2235 &  0.079687 &  0.723488 &  0.872963 &  0.902225 &  0.084312 &  \textbf{0.949388} &  \textcolor{lblue}{0.067800} &  0.760537 &  0.890038 &  0.898825 &  0.820337 &  0.416150 \\
    speech & 3686 & 400 &       1.6549 &  0.028663 &  0.017612 &  0.022413 &  \textcolor{lblue}{0.010150} &  0.032262 &  0.017612 &  \textbf{0.036213} &  0.023300 &  0.017612 &  0.013438 &  0.013438 &  0.028532 \\
   thyroid & 3772 &   6 &       2.4655 &  \textcolor{lblue}{0.086938} &  0.511787 &  \textbf{0.588962} &  0.229038 &  0.149550 &  0.401275 &  0.225013 &  0.214950 &  0.419962 &  0.400875 &  0.300363 &  0.150413 \\
 vertebral &  240 &   6 &      12.5000 &  0.043200 &  0.018537 &  0.054562 &  0.008925 &  0.061063 &  0.043200 &  0.066275 &  \textcolor{lblue}{0.000000} &  \textcolor{lblue}{0.000000} &  0.052125 &  0.052125 &  \textbf{0.096787} \\
    vowels & 1456 &  12 &       3.4341 &  0.345812 &  0.165612 &  0.212462 &  0.205875 &  0.396125 &  0.295687 &  0.341625 &  \textcolor{lblue}{0.010875} &  0.425825 &  0.380725 &  0.386688 &  \textbf{0.553763} \\
       wbc &  378 &  30 &       5.5556 &  0.545925 &  0.610650 &  0.565025 &  0.608113 &  0.559813 &  0.602900 &  0.508925 &  \textbf{0.673787} &  0.593287 &  0.602900 &  0.552400 &  \textcolor{lblue}{0.359000} \\
      wine &  129 &  13 &       7.7519 &  0.302675 &  0.383038 &  0.298513 &  0.221125 &  0.365175 &  \textcolor{lblue}{0.099413} &  0.314588 &  0.396425 &  0.365175 &  0.365175 &  0.365175 &  \textbf{0.497625} \\
\midrule
      AVG &  2149 &  58.8 &  11.4535 &  0.272119 &  0.409303 &  0.426580 &  0.347645 &  0.308189 &  0.408445 &  0.310041 &  0.390478 &  0.428982 &  0.437972 &  0.419610 &  0.375738 \\
\bottomrule
\end{tabular}
}
\end{table*}

Since our algorithm is based on kNN, it is slower than the algorithms with linear complexity.
Table \ref{table runtime} lists the run time of all algorithms on the three datasets with the largest number of data points.
In fact, the current Python implementation of our algorithm can be further optimized.

\begin{table*}[ht]
\setlength{\tabcolsep}{3pt}
\centering
\caption{Run time of each algorithm, maximum time highlighted in bold black and minimum time in light blue.}
\label{table runtime}
\vspace{0.1in}
\footnotesize
\scalebox{0.78}{
\begin{tabular}{llllllllllllllll}
\toprule
      \textbf{Data} &    $n$ &   $d$ & \textbf{Outlier\%} &        \textbf{FB} &      \textbf{HBOS} &   \textbf{IForest} &      \textbf{LODA} &       \textbf{LOF} &     \textbf{OCSVM} &      \textbf{LSCP} &     \textbf{COPOD} &       \textbf{kNN} &    \textbf{BikNN1} &    \textbf{BikNN2} &    \textbf{BikNN3} \\
\midrule
annthyroid & 7200 &   6 &       7.4167 &   1.02975 &  \textcolor{lblue}{0.00355} &  0.43840 &  0.05535 &  0.23535 &  0.60240 &   \textbf{6.07825} &  0.09580 &  0.43535 &   1.82260 &   1.53590 &   0.65075 \\
 satellite & 6435 &  36 &      31.6395 &   5.40105 &  \textcolor{lblue}{0.01300} &  0.64025 &  0.05285 &  0.78340 &  1.01280 &  \textbf{15.7053} &  0.19510 &  1.02475 &   3.24830 &   3.14705 &   2.33375 \\
satimage-2 & 5803 &  36 &       1.2235 &   3.68415 &  \textcolor{lblue}{0.01200} &  0.52010 &  0.05435 &  0.56000 &  0.83225 &  \textbf{13.9985} &  0.17450 &  0.78840 &   2.75765 &   2.77405 &   1.96275 \\
\bottomrule
\end{tabular}
}
\end{table*}

\section{Conclusions}

In this paper, we propose BikNN, a bilateral anomaly estimator based on k-Nearest Neighbors.
We measure the distances in density domain between each point and its k-Nearest Neighbors in spatial domain.
An anomaly coordinate system is built by collecting two unilateral anomalies from k-nearest neighbors of each point.
Then the anomaly score is estimated in 2D Anomaly Space.
Experiments performed on the synthetic and real world datasets demonstrate that our method performs well and achieve highest average performance.
We also show that our method can classify and visualize the anomalies of data points in 2D plane.
This kind of visualization tool can provide useful guidance for ones to further investigate what features make certain points potentially outliers.
One of our main contributions is that we present a framework based on k-Nearest Neighbors for anomaly estimation.
In future, we will explore more relationships between the neighboring points and integrate other powerful anomaly estimators into our framework.

\bibliographystyle{unsrt} 
\bibliography{references}  

\end{document}